\documentclass[a4paper,fleqn]{cas-dc}

\usepackage{soul}
\usepackage{color, xcolor} 
\usepackage[numbers]{natbib}
\usepackage{tabularx}

\begin{document}

\shorttitle{Large Language Models for Education: A Survey} 
\shortauthors{H. Xu \textit{et al.}}

\title [mode = title]{Large Language Models for Education: A Survey}   


\author[1]{Hanyi Xu}
\ead{xhyzhiyi@gmail.com}
\address[1]{College of Cyber Security, Jinan University, Guangzhou 510632, China}

\author[1]{Wensheng Gan}
\cortext[cor1]{Corresponding author}
\ead{wsgan001@gmail.com}
\cormark[1]

\author[2]{Zhenlian Qi}
\ead{qzlhit@gmail.com}
\address[2]{School of Information Engineering, Guangdong Eco-Engineering Polytechnic, Guangzhou 510520, China}
\cormark[1]

\author[1]{Jiayang Wu}
\ead{csjywu1@jnu.edu.cn}

\author[3]{Philip S. Yu}
\ead{psyu@uic.edu}
\address[3]{Department of Computer Science, University of Illinois Chicago, Chicago, USA}

\begin{abstract}
    Artificial intelligence (AI) has a profound impact on traditional education. In recent years, large language models (LLMs) have been increasingly used in various applications such as natural language processing, computer vision, speech recognition, and autonomous driving. LLMs have also been applied in many fields, including recommendation, finance, government, education, legal affairs, and finance. As powerful auxiliary tools, LLMs incorporate various technologies such as deep learning, pre-training, fine-tuning, and reinforcement learning. The use of LLMs for smart education (LLMEdu) has been a significant strategic direction for countries worldwide. While LLMs have shown great promise in improving teaching quality, changing education models, and modifying teacher roles, the technologies are still facing several challenges. In this paper, we conduct a systematic review of LLMEdu, focusing on current technologies, challenges, and future developments. We first summarize the current state of LLMEdu and then introduce the characteristics of LLMs and education, as well as the benefits of integrating LLMs into education. We also review the process of integrating LLMs into the education industry, as well as the introduction of related technologies. Finally, we discuss the challenges and problems faced by LLMEdu, as well as prospects for future optimization of LLMEdu.
\end{abstract}


\maketitle
\section{Introduction} \label{sec:Introduction}

\begin{figure}[ht]
    \centering
    \includegraphics[scale=0.16]{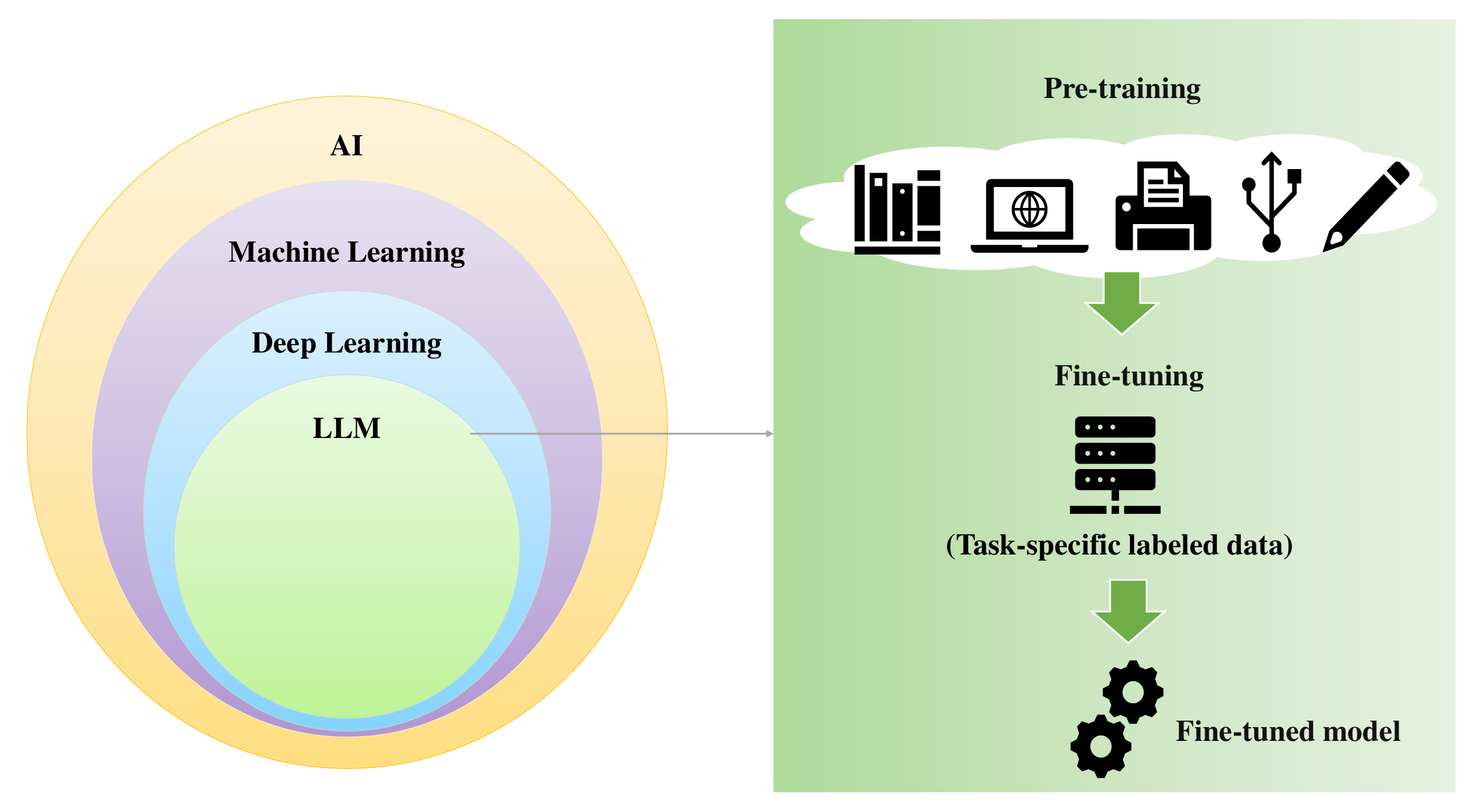}
    \caption{Framework of LLMs.}
    \label{fig:stru}
\end{figure}

Artificial intelligence (AI) has developed rapidly in recent years \cite{lv2020trustworthiness,wang2020artificial,zhao2020overview}, thanks to the continuous improvements in Web 3.0 \cite{gan2023web}, Internet of Behaviors (IoB) \cite{sun2023internet}, data mining \cite{gan2021survey,huang2023us,lin2015mining}, deep learning \cite{xie2021deep}, and language processing technologies \cite{hu2023survey}. LLMs have shown excellent performance in various industries with the optimization of pre-training models and the continuous adjustment of related technologies \cite{chung2021w2v,zeng2023distributed}. LLM is mainly based on many AI technologies, e.g., natural language processing (NLP), and was used to understand and generate massive texts \cite{gu2021domain}. They perform self-supervised learning on a large-scale corpus to obtain the statistical laws of language \cite{elnaggar2021prottrans} and then convert it into logical natural language text. Its basic framework is shown in Figure \ref{fig:stru}. LLMs have demonstrated strong versatility and logical reasoning capabilities, leading to their widespread model-as-a-service (MaaS) \cite{gan2023model} in various industries, including finance, education \cite{gan2023large}, law \cite{lai2023large}, robotics \cite{zeng2023large}, and government affairs \cite{chang2023survey,fan2023recommender,yan2023practical}. Creating a scenario-based user experience is a key advantage for most digital companies, and it also happens to be a development need for LLM.

The concept of education has been around for centuries, dating back to the theory of biological origins. In primitive societies, education was limited to the use of primary production tools, whereas ancient societies relied on oral transmission and practice to pass knowledge down to future generations \cite{lin2022metaverse}. With the development of science and technology in modern society, education and AI have become inseparable \cite{chen2020multi}, including intelligent teacher assistants, voice assistants \cite{malodia2021people,qidwai2020humanoid}, AI writing creation platforms, etc. The fourth industrial revolution, represented by the intelligent revolution \cite{brem2021ai}, can bring the education industry to a new level with the help of LLMs. Education is essentially about knowledge transfer, instant feedback, and emotional interaction. LLMs mainly enhance the ``immediate feedback" process in education. They have the potential to revolutionize the education industry by providing personalized, adaptive learning experiences for students. By infusing knowledge into their models, LLMs can gradually build a deep understanding of the world, surpassing human learning in some aspects. They can generate high-quality text content, comprehend natural language, extract information, and answer questions across various fields \cite{liu2023summary}. LLMs can also do complex mathematical reasoning \cite{Xu2023}, which helps the education sector show that they are good at self-supervision, intelligent adaptive teaching, and multi-modal interaction \cite{Deng_2023_CVPR}. With their ability to adapt the individual students' needs and learning styles, LLMs can provide a more effective and engaging learning experience.

\textbf{Research gaps}: There are already many educators and researchers who have shown a lot of thinking about AI in education. Examples are as follows: Some research has been conducted on the paradigm shift in AI in education \cite{OUYANG2021100020} and on the impact of AI in management, teaching, and learning \cite{chen2020artificial}. Some studies explain AI in education and show how they work \cite{luckin2016intelligence}. Due to the rapid iteration and update of AI, many new educational AI technologies have been spawned, but there is a lack of summary and analysis of emerging technological means. LLMs, as one of these technologies, have significantly advanced AI development to a new stage. LLMs are the latest technological means to support intelligent education. The integration of education and LLMs particularly highlights the development and application characteristics of LLMs. There has been one brief review of LLMs for education \cite{gan2023large}, while many characteristics of LLMEdu and key technologies are not discussed in detail.

\textbf{Contributions}: To examine the potential of LLMEdu and promote its development, this paper provides an in-depth analysis of the development process and technical structure of LLMEdu and forms a comprehensive summary. This review aims to help readers gain a deeper understanding of LLMEdu and encourages us to invent and consider LLMEdu applications. The specific contributions are as follows:

\begin{itemize}
    \item We take a closer look at the connection between LLMs and education, aiming to achieve smart education.
    \item We demonstrate the development process of LLMEdu through the process of applying LLMs to education and the key technologies of LLMs.
    \item We review the implementation of LLMEdu from the perspective of LLMs empowering education, focusing on exploring the development potential of LLMEdu.
    \item We highlight the problems and challenges existing in LLMEdu in detail, aiming to trigger some insight, critical thinking, and exploration.
\end{itemize}

\textbf{Roadmap}: In Section \ref{sec:characteristics}, we briefly introduce the characteristics of LLMs and the education industry, as well as the characteristics of LLMs integrated into education. In Section \ref{sec:ilm}, we conduct an in-depth analysis of the process of applying LLMs to education. In Section \ref{sec:ktfe}, we explain the key technologies related to LLMs. In Section \ref{sec:cases}, we provide the implementation of LLMEdu from the perspective of empowering education with LLMs. In Section \ref{sec:challenges}, we highlight some of the main issues and challenges in LLMEdu. Finally, in Section \ref{sec:conclusion}, we summarize LLMEdu and propose expectations for the development of future LLMs. Table \ref{Symbols} describes some basic symbols in this article. 

\begin{table}[ht]
    \small
    \centering
    \caption{Summary of symbols and their explanations}
    \label{Symbols}
    \begin{tabular}{|c|l|}
        \hline
        \textbf{Symbol} & \textbf{Definition}    \\ \hline
        AI  & Artificial Intelligence\\ \hline
        AIGC & AI-Generated Content\\ \hline
        ChatGPT & Chat Generative Pre-Training Transformer \\ \hline
        CV & Computer Vision\\ \hline
        DNNs & Deep Neural Networks\\ \hline
        GPT & Generative Pre-trained Transformer\\ \hline
        HFRL & Human Feedback Reinforcement Learning \\ \hline
        LLMEdu & Large Language Models for Education\\ \hline
        LLMs &  Large Language Models\\ \hline
        LMs & Language Models\\ \hline  
        NLP & Natural Language Processing\\ \hline
    \end{tabular}
\end{table}

\section{Characteristics of LLM in Education}
\label{sec:characteristics}

\begin{figure}[ht]
    \centering
    \includegraphics[scale=0.17]{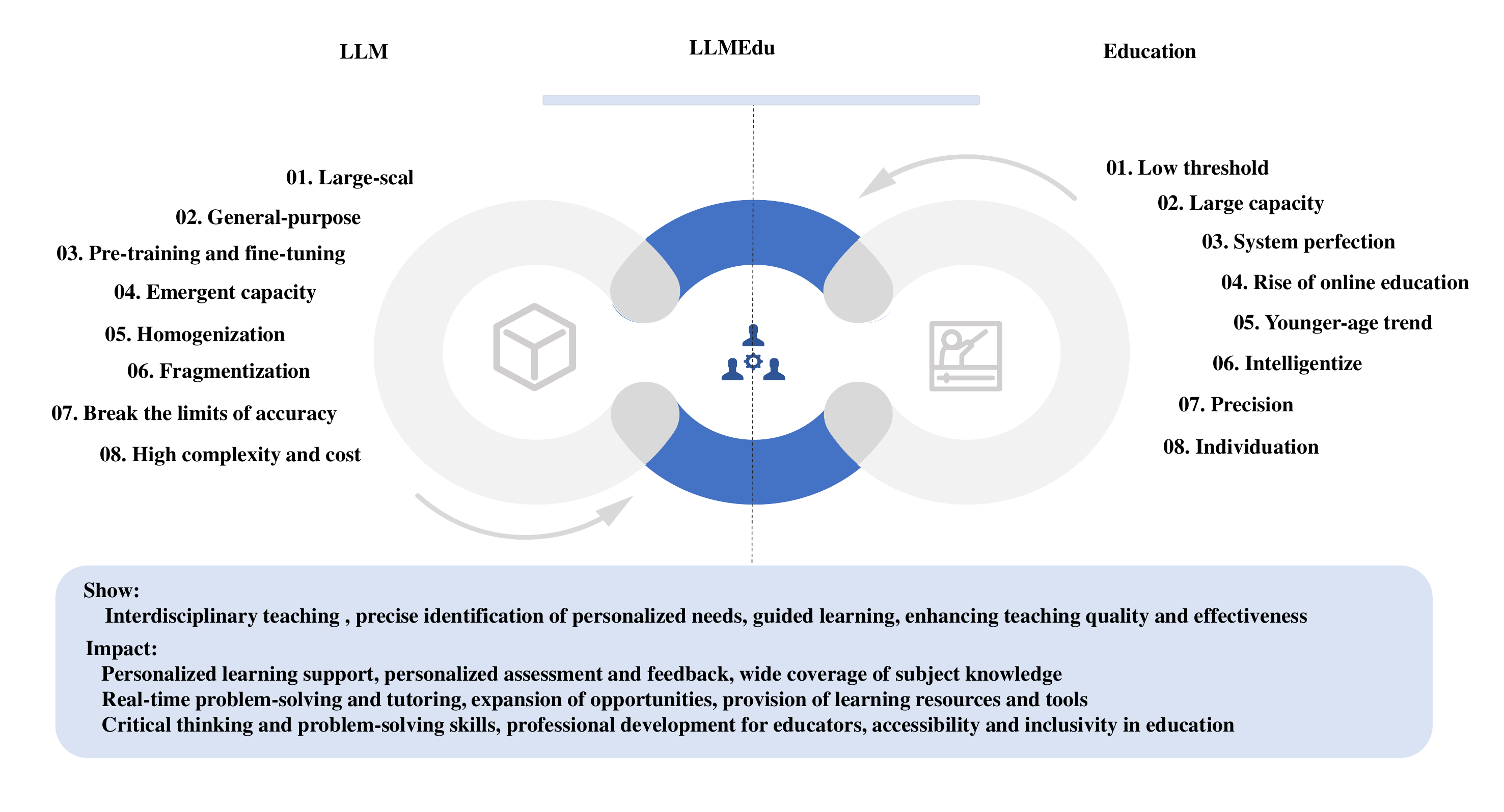}
    \caption{The characteristics of LLMEdu.}
    \label{fig:cha}
\end{figure}

In this section, we discuss the key characteristics of LLMs, the key characteristics of education, the limitations of traditional education, and the combinations between LLMs and education, as depicted in Figure \ref{fig:cha}.

\subsection{Characteristics of LLMs}

\textbf{Large-scale}. The term 
``large" in LLMs can be interpreted in two ways. Firstly, LLMs possess an enormous number of parameters, with the parameter count increasing exponentially from billions to trillions in just a few years. For instance, Google's BERT had 300 million parameters in 2018, GPT-2 had 1.5 billion parameters in 2019, and GPT-3 had 175 billion parameters in 2021 \cite{zhang2021commentary,2022arXiv220203480S}. In 2022, the Switch Transformer reached an impressive 1.6 trillion parameters \cite{lin2021m6,schwartz2020green}. Furthermore, LLMs are trained on vast amounts of data from diverse sources, including the web, academic literature, and conversations. This large-scale corpus of data enables the models to learn and represent complex patterns and relationships in language, leading to improved performance in various NLP tasks \cite{thirunavukarasu2023large}.

\textbf{General-purpose}. LLMs have a wide range of applications \cite{pan2023unifying}. In addition to excelling in specific domains, they are adept at handling various types of tasks, including NLP, CV, speech recognition, and even cross-modal tasks. In other words, LLMs possess powerful generalization capabilities, and achieving such capabilities requires training on massive amounts of data.

\textbf{Pre-training and fine-tuning} \cite{derose2020attention,hu2023survey,zeng2023distributed}. The core of the model training process lies in the use of pre-training followed by fine-tuning. Initially, pre-training is performed on a large-scale unlabeled text corpus to acquire the model's basic language knowledge. Subsequently, fine-tuning is conducted on specific tasks in a particular domain to better understand and generate language specific to that domain, such as legal, educational, or medical texts.

\textbf{Emergent ability: unpredictability} \cite{pan2023unifying}. The emergent ability of LLMs refers to their capacity to generate coherent and logically consistent text without explicit human intervention, as they have learned from their training process. When the amount of data reaches a sufficiently large scale, the model's learning and feedback capabilities can experience a substantial increase, resulting in improved performance.

\textbf{Fragmentation} \cite{rajbhandari2020zero}. The current AI landscape is characterized by diverse business scenarios across various industries, resulting in fragmented and diversified AI demands. The development process of AI models involves several stages, including development, hyperparameter tuning, optimization, and iterative deployment for eventual application. Each stage requires significant investment, and in high-cost situations, catering to customized market demands can be challenging.

\textbf{Potential for breaking accuracy limitations}. The development of deep learning has taken a long time. The improvement in accuracy through architectural changes appears to have reached a bottleneck as neural network design techniques have matured and converged. However, LLM development has shown that increasing the scale of both the model and the data can help break through accuracy limitations. Research experiments have consistently demonstrated that scaling up the model and data leads to improved model accuracy \cite{tan2019efficientnet}.
\textbf{High complexity and investment costs}. LLMs are becoming increasingly complex, with single-step computation time growing by more than 10 times \cite{bai2021}. For high-traffic businesses, a training experiment that used to take a few hours now takes several days, with the expectation that tests will remain within a one-day timeframe as a basic requirement \cite{ma2023llm}. Moreover, training a general-purpose large model is expensive, and if subsequent optimization, updates, and deployment are included, it will cost even more. For example, the core infrastructure of ChatGPT, the Azure AI, required an investment of nearly \$1 billion \cite{D2020}. Moreover, ChatGPT has high requirements for the number of GPU chips used for data processing \cite{narayanan2021efficient}.

\subsection{Characteristics of education}

According to its definition, education is a deliberate and conscious social practice that aims to nurture individuals. Its fundamental characteristic is its process-oriented nature, indicating that education exists and evolves through a series of steps. With a focus on individuals, education ultimately aims to facilitate their holistic and enduring growth. Education encompasses knowledge transmission, immediate feedback, and emotional interaction. Error correction, knowledge reinforcement, and rapid training consolidation are some parts of educational behavior. Furthermore, the education system is highly intricate, marked by the distinctiveness of its subjects, diverse requirements, and intricate interactions.

\subsubsection{Educational development process}

\textbf{Low entry barriers.} On one hand, the accessibility of starting an educational institution is relatively easy \cite{budiharso2020improving}, resulting in lower operating and investment costs for both teachers and institutions. However, this has also led to a disparity in teacher qualifications, contributing to issues such as disorder in the education and training industry, misleading advertisements, exaggerated titles for teachers, and ineffective offline one-on-one teaching. These have subsequently led to an increase in complaints. On the other hand, there has been a reduction in barriers to education for learners, leading to greater equality of educational opportunities across different regions and a stronger emphasis on the right to education.

\textbf{Large capacity \cite{li2020education}.} The education industry encompasses a significant number of students and teachers, making it crucial to consider the implications of a large population. Moreover, there exists a diverse array of educational settings, including public schools as well as numerous private educational institutions. There is an abundance of educational materials available, and the advent of the internet has made access to educational resources easier. This development has transcended the confines of traditional textbook-based teaching, breaking down information barriers and expanding the horizons of education.

\textbf{Well-developed system.} The expansion of education has been propelled by economic development \cite{kopnina2020education}, leading to a surge in investment in the education sector. This growth encompasses a wide range of educational institutions at different levels. Moreover, the education system encompasses diverse forms of education, such as social life education, family education, and school education. It also encompasses a variety of disciplines, including mathematics, languages, and physical education.

\textbf{Rise of online education \cite{koksal2020rise}. } Since the late 1990s, emerging technologies have made significant inroads into the education industry \cite{bunnell2020british}. This transformation has propelled education through various stages, including traditional education, digital education, internet-based education, mobile-based education, and intelligent education. The advancement of information technology has played a pivotal role in facilitating education development by overcoming time and space constraints, making knowledge acquisition more convenient and rapid.

\textbf{Education at a younger age.} The development of the internet has dismantled barriers to education, resulting in heightened parental concerns and an increased focus on early education. Under the influence of globalization, the significance of early education \cite{yang2019changing}, particularly in language and logic development, has been recognized. In conjunction with the surge of online education, early childhood education has become more readily available. A wide range of tutoring classes and early learning programs have become commonplace.

\textbf{Intelligent, precise, and personalized education \cite{chen2020application}.} With the rapid advancement of AI, technology has significantly enhanced production methods and raised people's living standards. As a result, society's demand for education has escalated, leading to a more targeted approach to talent development. Education is currently transforming the integration and innovation of ``AI + education" in smart education.

Although education has integrated AI to a significant extent, the nature of human education and machine education fundamentally differs in a two-tier manner. These two forms of education vary in their sequence: human education primarily focuses on shaping values, followed by systematic knowledge acquisition, and ultimately engaging in real-world experiences to foster learning. In contrast, machine education begins by processing vast amounts of data, subsequently discerning between right and wrong (learning values), incorporating human feedback, and ultimately attaining practicality. When it comes to learning, the most notable distinction between humans and machines lies in the limited energy humans possess to acquire knowledge within a fixed period, whereas machines have a relatively unlimited learning capacity. Embracing AI, formulating education strategies that align with the current era, and achieving a comprehensive digital transformation of education are the central points of contemporary educational development.

\subsubsection{Impact on teachers}

\textbf{Instructional method's development}. Digital education provides a wider range of teaching methods and tools \cite{dillenbourg2016evolution}. It requires teachers to adapt and become proficient in utilizing these innovative approaches and technologies. This includes leveraging online learning platforms, educational applications, and virtual classrooms to effectively impart knowledge and engage with students. To cater to student's diverse learning needs, teachers must acquire familiarity with and expertise in using these technologies.

\textbf{Personalized and self-directed learning support}. Digital education has the potential to better support personalized and self-directed learning \cite{butcher2011self}. Teachers can leverage technology to gain insights into student's learning styles, interests, and needs. They also provide tailored instructional content and learning plans. This shift in education will see teachers adopt more of a guide and mentor role. They encourage students to take an active role in their learning and self-development.

\textbf{Data-driven instructional decision-making}. Digital education yields a wealth of learning data, including student's performance, interests, and progress \cite{zhao2022study}. Teachers can leverage this data to make informed instructional decisions and provide personalized guidance. By analyzing student's data, teachers can identify areas of difficulty and weakness and offer targeted support and feedback to help students overcome these challenges and improve their learning outcomes.

\textbf{Collaboration and cross-border teaching}. Digital education has the power to break down geographical barriers, enabling teachers to engage in cross-border teaching and collaboration with students from all over the world. This allows for the sharing of instructional resources, experiences, and best practices among educators, promoting professional development and collaboration within the teaching community.

\textbf{Cultivating 21st-century skills}. In the digital age, it's essential for students to develop skills such as creative thinking, digital literacy, collaboration, and problem-solving \cite{hsu2019developing}. Teachers play a vital role in guiding students to cultivate these skills and providing relevant educational support and guidance. By exploring and applying new technologies together with students, teachers can foster student's innovation and adaptability, preparing them for success in an ever-changing digital landscape.

Teachers are indispensable in the digital transformation of education, as they play a multifaceted role in shaping student's academic, emotional, and social development. While technology can provide access to vast knowledge and resources, it cannot replace the personalized guidance, emotional support, and values-based education that teachers offer. The expertise, interpersonal relationships, and educational wisdom of teachers are still essential elements in the digital transformation of education, ensuring that students receive a well-rounded education that prepares them for success in the 21st century.

\subsubsection{Educational challenges}

\textbf{Personalized learning needs. } In contemporary education, students have diverse learning needs, styles, interests, and aspirations. The traditional one-size-fits-all approach may not cater to each student's unique requirements, and personalized learning is essential to addressing these differences effectively. Therefore, implementing personalized learning is a significant challenge that educators and administrators must address to ensure that every student receives an education tailored to their individual needs and abilities.

\textbf{Insufficient educational resources.} Despite the advancements in technology, there are still areas where schools lack modern technology infrastructure, resulting in a digital divide that hinders student's access to online learning and digital education resources. Moreover, the number of students worldwide continues to rise, putting immense pressure on the education industry. Some regions face the challenge of insufficient educational resources, including teachers, classrooms, and learning materials, leading to disparities in educational opportunities.

\textbf{Education quality and standards.} Inconsistencies in education quality pose a significant challenge. In some regions, an exam-oriented approach to education may lead to a narrow focus on standardized testing, resulting in a simplified curriculum and a lack of support for students' personal interests and development. Ensuring high-quality, standardized education is crucial to enhance student's academic performance and overall quality. This can be achieved by implementing a well-rounded curriculum that fosters critical thinking, creativity, and problem-solving skills while also providing individualized support for student's unique needs and interests.

\textbf{Diverse educational technology.} The integration of big data, AI, virtual reality, and other educational technologies has the potential to revolutionize the education sector. However, it also poses new challenges, such as management, security, and privacy considerations. Effective integration and utilization of these technologies are crucial to enhance the learning experience and achieve optimal educational outcomes. This requires a well-thought-out strategy that takes into account the unique needs and constraints of the education sector.

\textbf{Challenges in implementing new educational concepts.} The rapid pace of technological and economic advancements, coupled with improvements in living standards and quality, has led to the emergence of new educational concepts. One such concept is ``Science Technology Engineer Art Math (STEAM)" education, which emphasizes interdisciplinary approaches and hands-on practice. However, implementing these cutting-edge educational concepts and cultivating the next generation of socially conscious talents pose a significant challenge for the education sector. Effective strategies and innovative approaches are needed to address these challenges and ensure that students are well-equipped to thrive in an ever-changing world.

\subsection{Characteristics of LLMEdu}

The integration of AI into the education industry has accelerated rapidly \cite{gao2020personalized,li2006restructuring,tang2020personalized}, transforming teaching methods and enhancing learning outcomes. From computer-assisted teaching to personalized adaptive learning and content generation, AI has revolutionized the education sector, catering to diverse age groups and fields of study. In the era of intelligence, the primary objective of education is to convert knowledge into intelligence and nurture intelligent individuals. LLMs, with natural language technology at their core, align seamlessly with the education industry's development and adapt to the vast changes in intelligent education. These models have the potential to support and enhance various aspects of the learning experience, making education more accessible, engaging, and effective.

\subsubsection{Specific embodiment of ``LLMs + education"}

\begin{figure}[ht]
    \centering
    \includegraphics[scale=0.16]{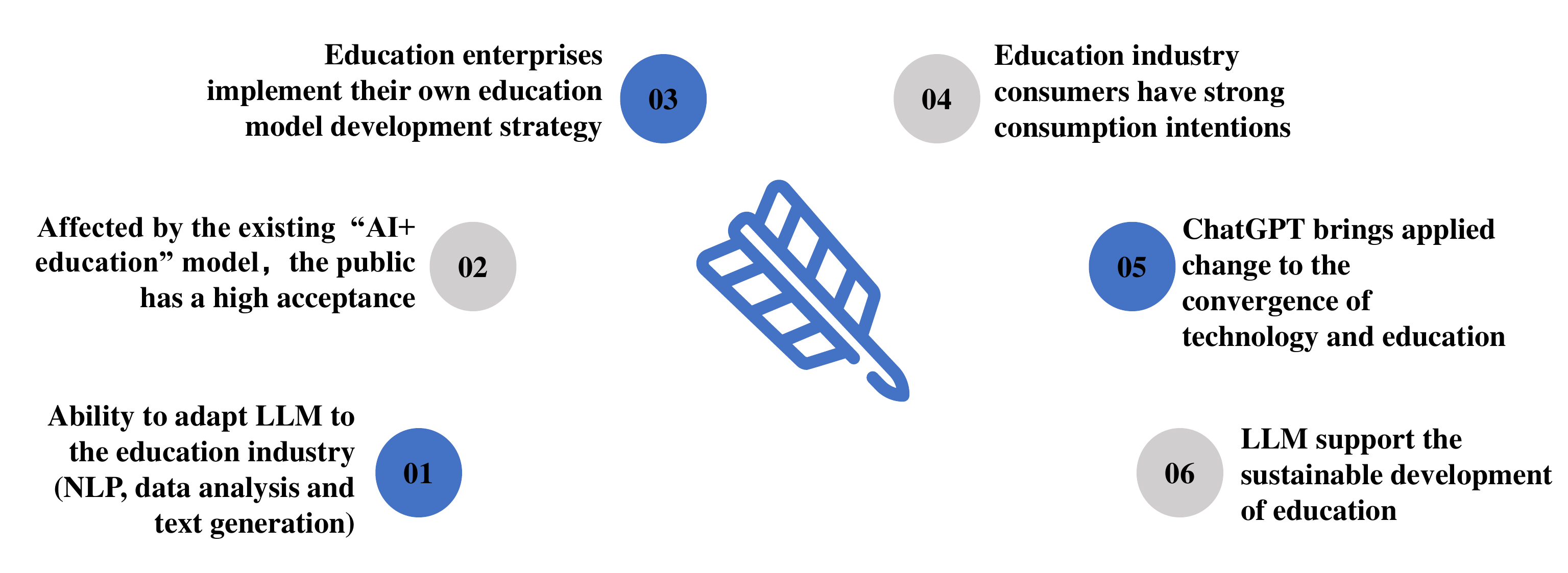}
    \caption{Reasons for integrating LLM into education.}
    \label{fig:reason}
\end{figure}

Reasons for integrating LLM into education are shown in Figure \ref{fig:reason}.

\textbf{Interdisciplinary teaching \cite{lyu2023new}.} The training of LLMs with vast amounts of data gives them a significant advantage in knowledge integration. They can provide diverse learning support based on different subjects and boast excellent interdisciplinary capabilities. For instance, the ``Ziyue" large model\footnote{\href{https://www.bing.com/ck/a?!&&p=684b97a33aeec65fJmltdHM9MTY5OTE0MjQwMCZpZ3VpZD0xODBlMmVkZC1jNGM5LTYxYWEtMGNmYi0zZGMxYzUxYjYwN2YmaW5zaWQ9NTIwOQ&ptn=3&hsh=3&fclid=180e2edd-c4c9-61aa-0cfb-3dc1c51b607f&psq=\%e5\%ad\%90\%e6\%9b\%b0\%e6\%95\%99\%e8\%82\%b2\%e5\%a4\%a7\%e6\%a8\%a1\%e5\%9e\%8b\%e8\%8b\%b1\%e6\%96\%87\%e5\%90\%8d&u=a1aHR0cHM6Ly9haWNlbnRlci55b3VkYW8uY29tLw&ntb=1}{https://aicenter.youdao.com} } prioritizes a ``scenario-first" approach, while the iFLYTEK ``Spark Desk"\footnote{\href{https://passport.xfyun.cn/login}{ (xfyun.cn)} } can conduct human-like interactive learning in various fields, including mathematics, English oral practice, essay correction, and more. These models have the potential to revolutionize the way we learn and teach \cite{cheng2023artificial}.

\textbf{Precise identification of personalized needs.} LLMs possess advanced language understanding and generation capabilities, enabling them to provide adaptive learning guidance tailored to individual users' age, learning stage, and learning environment. For example, the iFlytek learning machine based on LLMs can provide customized teaching for traditional subjects, such as oral teaching, Chinese and English composition correction, interactive supplementary mathematics, and so on, providing students with personalized one-to-one mentoring experiences. Furthermore, the learning machine can help parents answer questions through one-to-one dialogue, provide suggestions, and assist in parent-child communication, parent-child interaction, behavioral habits, and so on.

\textbf{Guided learning.} LLMs are shifting towards a more human-like approach, providing authentic conversational teaching experiences in various scenarios instead of simply giving answers. This is particularly noticeable in subjects like physics and mathematics, where LLMs simulate a teacher's role and ask questions to encourage critical thinking and independent exploration \cite{jeon2023large}. By fostering a self-learning environment, LLMs can help students develop their problem-solving skills and become more effective learners \cite{mhlanga2023open}. For example, OpenAI collaborated with the educational organization Khan Academy to produce Khanmigo, an LLM-based educational tool. As students complete the exercises, Khanmigo can guide them to get answers on their own by asking a lot of questions.

\textbf{Integration of three modes.} Tool-based, companion-based, and information-based \cite{edyko2023utilizing,ivanov2023game,williamson2023re}. The tool-based mode primarily involves using data to construct a knowledge base, which becomes a large-scale query repository. The companion-based mode is exemplified by virtual teachers and assistants, providing virtual teaching and online assistance through human-like conversations. The informatization-based mode mainly refers to educational informatization, accelerating the development of an ``internet + education" platform.

\subsubsection{Impact of ``LLMs + education"}
``LLMs + education" will have far-reaching and profound impacts. Here are 10 areas where these impacts can be observed, along with detailed explanations.

\textbf{Personalized learning support.} LLMs can provide customized learning support based on students' personalized needs. By deeply understanding students learning characteristics, interests, and learning styles, LLMs can tailor teaching content and learning plans for each student. For example, in mathematics learning, LLMs can provide targeted guidance for students' weak points in mathematics by interacting with them in dialogue, helping them overcome difficulties, and improving their mathematical abilities. LLMs can design adaptive tests that adjust the difficulty of questions based on students’ responses, accurately assessing students’ knowledge levels and ensuring they are educated at the appropriate level \cite{ahmad2023generative}.

\textbf{Personalized assessment and feedback.} LLMs can provide personalized assessment and feedback based on students' learning performance \cite{latif2023artificial}. By analyzing student's answers, understanding levels, and error patterns during the learning process, LLMs can provide targeted assessment results and improvement suggestions. For example, when students encounter difficulties in writing, LLMs can analyze the structure, grammar, and expression of their writing pieces and provide detailed guidance and suggestions to help students improve their writing skills \cite{al2023chatgpt,maddigan2023chat2vis}. Some commercial auxiliary tools based on OpenAI's LLM technology, MagicSchool, and Eduaide, can participate in the assessment of students' homework and give feedback \cite{pankiewicz2023large}.

\textbf{Wide coverage of subject knowledge.} LLMs have extensive knowledge coverage and can encompass knowledge content from multiple subject areas \cite{liu2023m3ke}. Students can engage in dialogue with LLMs to acquire knowledge and information across various subject domains. For instance, when students encounter problems in history learning, LLMs can provide detailed explanations and in-depth discussions of historical events, figures, and backgrounds, helping students better understand historical knowledge. According to statistics, the latest model has 13 trillion tokens of carefully selected pre-training knowledge data, which is equivalent to 5 million sets of four major classics. In addition, 1.8 trillion "knowledge fragments" are extracted during training \cite{borgeaud2022improving}.

\textbf{Interdisciplinary learning.} LLMs have excellent interdisciplinary capabilities, enabling students to engage in integrated learning and cultivate interdisciplinary thinking skills \cite{valverde2023interdisciplinary}. Through interactions with LLMs, students can integrate and apply knowledge from different subject areas. For example, when conducting scientific experiments, students can have conversations with LLMs to discuss experimental principles, data analysis, and scientific reasoning, promoting integrated learning between science and mathematics, logical thinking, and other disciplines \cite{amer2023large}.

\textbf{Real-time problem-solving and tutoring.} LLMs can provide real-time problem-solving and tutoring support for students. When students encounter confusion or questions during the learning process, they can ask LLMs at any time and receive immediate answers and solutions. A survey report in the first half of this year pointed out that 89\% of American students surveyed were using ChatGPT to complete homework \cite{zeng2023chatgpt}. Additionally, when students encounter comprehension difficulties while reading literary works, they can engage in dialogue with LLMs to explore the themes, plots, and character images of literary works, helping students better understand and analyze literary works \cite{wang2023enhancing}.

\textbf{Opportunities for learning across time and space.} The existence of LLMs allows students to learn anytime and anywhere. Students can interact with LLMs through mobile devices or computers, without being constrained by traditional classroom time and location. For example, students can utilize evening or weekend time to engage in online learning with LLMs, improving their academic abilities and knowledge levels. Online learning platforms, which utilize LLMs, provide students with access to a wide range of courses and disciplines via the Internet. The LLMs support the implementation of virtual classrooms and distance education, and students talk to the LLMs in real time to solve problems. 

\textbf{Provision of learning resources and tools.} LLMs can serve as rich learning resources and tools, providing a wide range of educational materials and tools for student's learning needs. For instance, LLMs can offer textbooks, educational videos, interactive exercises, and other learning materials to support student's learning in various subjects \cite{baidoo2023education}. Additionally, there are some subject-specific tools, such as MathGPT. MathGPT has an accuracy rate of 60.34\% in the benchmark test AGIEval, which can help students solve mathematical problems efficiently \cite{zhong2023agieval}.

\textbf{Promotion of critical thinking.} LLMs can guide students in developing critical thinking and problem-solving skills \cite{huang2022towards}. By engaging in dialogue and posing thought-provoking questions, LLMs can foster a thinking atmosphere that encourages students to explore answers, enhancing their self-learning abilities and critical thinking skills. For example, LLMs can simulate a teacher's role in a physics class, asking students questions about concepts, principles, and problem-solving strategies, encouraging them to think critically and develop problem-solving skills \cite{wang2020minilm}.

\textbf{Professional development for educators.} LLMs can support the professional development of educators by providing them with access to a vast amount of educational resources, best practices, and innovative teaching approaches. Educators can interact with LLMs to enhance their teaching methods and explore new ways to engage students \cite{lim2023sign}. For example, teachers can engage in dialogue with LLMs to discuss teaching strategies, classroom management techniques, and approaches to address student's individual needs, improving their teaching effectiveness and professional growth.

\textbf{Accessibility and inclusivity in education.} LLMs can contribute to making education more accessible and inclusive. They can provide learning support for students with different learning styles, abilities, and backgrounds, ensuring that all students have equitable access to quality education. For example, LLMs can offer alternative explanations, visual aids, and interactive learning experiences to accommodate diverse learners, including students with learning disabilities or language barriers, making education more inclusive and supportive. Additionally, through multicultural training, LLMs can better understand and respect students from different cultural backgrounds and create a learning environment that is inclusive and respectful of diversity.

In summary, the integration of LLMs with education will revolutionize the learning experience by providing personalized support, expanding knowledge coverage, promoting critical thinking, and enhancing the accessibility and inclusivity of education. It will empower students and educators alike, transforming the way knowledge is acquired, shared, and applied in the digital age.

\section{How to Gradually Integrate LLMs into Education}  \label{sec:ilm}

The integration of AI into the education industry has been progressing step by step, from machine learning (implementing the ability to store and calculate) to deep learning (implementing the ability to see and hear), and now to LLMs (capable of understanding and creating) \cite{meng2020text,schlexer2019machine,wang2020survey}. In the current era, the vigorous development of quality education by the entire population and the active deployment of educational intelligent hardware nationwide represent the active transformation of educational training enterprises \cite{biggs2022ebook,philippe2020multimodal}. In the long-standing coexistence and collaboration between teachers and AI models \cite{wang2019human}, as well as the highly homogeneous hardware background, LLMs have emerged as one of the most important technologies in human intelligence.

\subsection{Reasons why LLMs for education}

LLMs' excellent characteristics make their application in the education industry very reasonable. NLP \cite{gu2021domain}, data analysis \cite{gan2022discovering,zhang2022tusq}, and text generation capabilities \cite{wu2023ai} align well with the fundamental processes of learning, questioning, and feedback in education. The iterative optimization process of ``development-deployment" suits the application process in the education industry. User testing and feedback data lay the foundation for further optimization. Taking the development of LLMs in China as an example, the Spark Desk by iFLYTEK\footnote{\href{https://xinghuo.xfyun.cn/}{https://xinghuo.xfyun.cn/}}, the ERNIE Bot by Baidu\footnote{\href{https://yiyan.baidu.com/}{baidu.com} }, and the ``MathGPT" by TAL\footnote{\href{https://math-gpt.org/}{MathGPT | AI Math Calculator (math-gpt.org)} } have accumulated data from years of experience in the education industry \cite{zou2021pre}. During their usage, these LLMs can collect more data from the education industry, leading to further technology optimization.

The ``AI + education" model has already formed, and the gradual maturity of AI technology has paved the way for the entry of LLMs into the education industry. Smart classrooms, voice-assisted teaching, intelligent problem-solving, and other AI applications have become routine in the education industry, leading to high acceptance of LLMs \cite{beck1996applications,bhutoria2022personalized,saini2019smart}. It is important to recognize that LLMs are the latest technological achievements that gather human collective intelligence, rather than only technological achievements. However, LLMs' development potential and influence are gradually increasing.

Education companies implement their own LLMEdu development strategies. LLMs require massive amounts of data and significant investments to support them. In terms of data, looking at various education companies, long-term experience data accumulation, technology accumulation, and an objective combination of their development conditions have differentiated the educational application of LLMs. They focus on LLM research and strive to maximize their benefits, cater to current development trends, and reduce development costs. In terms of funding, consumers in the education industry have a strong willingness to consume. As people's living standards and education levels improve, the world strengthens the education industry and injects large amounts of funding to provide a solid foundation for LLM research, development, and application.

ChatGPT makes practical changes to the integration of technology and education. Learning is an exploration process, and LLMs play an exploratory role in education. Because of interactive questions and answers, people's roles are changing from passive recipients of knowledge to active explorers. Because of the existence of machine hallucinations, scholars need to have a skeptical and judgmental attitude towards generated knowledge and treat LLMs from a dialectical perspective. Intelligent technology stimulates human creativity, allowing people to continuously expand their breadth of learning, thus leading to scientific and technological progress.

LLMs support the sustainable development of education \cite{bahrami2023examining}. Innovation is the core of technological development and the premise of long-term application. By fully utilizing AI technologies such as ChatGPT, the application process in education can transition from a search mode to a content generation mode personalized for individuals. This enables the development of diverse, scalable, tangible application scenarios, as well as a series of differentiated and highly experiential educational products and services. It provides excellent environments and resources for educators and education recipients, supporting education's sustainable development.

Nowadays, general language models (LMs) leverage extensive data memory to shift from dedicated to universal application models. They rely on text generation capabilities, transitioning the application process from distribution to generation. This allows them to achieve multi-modality and transform application scenarios from single to multiple \cite{han2023imagebind}. Multi-modal LLMs, which combine pre-training and downstream tasks, can efficiently complete downstream task adaptation with relatively small amounts of data and can be used in small sample learning and natural language question answering. In education, three typical applications are realized: automatic generation of teaching resources, human-machine collaborative process support \cite{zhipeng2019jiuge}, and intelligent teaching assistance for teachers. Multi-modal LMs combine the three fields of reinforcement learning, CV, and NLP. They attempt to extend the concept of LMs \cite{huang2023chatgpt,rudovic2019multi,tao2021multi}.

What's more, we demonstrate the development of the GPT models, as shown in Table \ref{Ic}.

\begin{table*}[ht]
\renewcommand{\arraystretch}{1.8}
    \centering
    \caption{Iteration and comparison of LLMs}
    \label{Ic} \resizebox{0.9\linewidth}{!}{ 
    \begin{tabular}{|c|c|c|c|c|c|}
        \hline
       \textbf{LLMs}& \textbf{Publish time}& \textbf{Parameter quantity} & \textbf{Pre-training data size} & \textbf{Training paradigm} & \textbf{Feature} \\ \hline
        GPT & 2018.7 & 120 million & 5G & Pre-training + fine-tuning & Reflection of the advantages of self-attention structure \\ \hline
        
        GPT-2\footnote{https://openai.com/research/gpt-2-1-5b-release} & 2019.2 & 1.5 billion & 40G & Prompt paradigm based on Tunning-free: Zero Shot Prompt & Open the exploration of the Prompt paradigm\\ \hline
        
        GPT-3\footnote{https://openai.com/blog/gpt-3-apps} & 2020.6 & 175 billion & 45TB & Prompt paradigm based on Tunning-free: In-Context Learning & Deepen the exploration of the Prompt paradigm\\ \hline
        
        InstructGPT\footnote{https://openai.com/research/instruction-following} & 2022.3 & 175 billion & 45TB & Prompt paradigm of Instruction Tuning & Start paying attention to human preferences \\ \hline
        	
        ChatGPT\footnote{https://openai.com/chatgpt} & 2022.11 & 175 billion & 45TB & Reinforcement learning from human feedback  & Aligned with human preferences \\ \hline
        
        GPT-4\footnote{https://openai.com/gpt-4} & 2023.3 & Nearly 2 trillion & - & Reinforcement learning from human feedback & Multimodal processing and getting closer to the bionic human brain\\ \hline
        
       LaMDA\footnote{https://openai.com/gpt-4https://blog.google/technology/ai/lamda/} & 2021 & 137 billion &  150TB & Pre-training + fine-tuning & Introduce external information retrieval system\\ \hline
        
       BARD\footnote{https://bard.google.com/} & 2023.2 & 137 billion & - & Join ChromeOS as a search engine & Using LaMDA as a base\\ \hline
         
       PaLM & 2022.4 & 540 billion & - & PathWay distributed training framework & Large scale, multi-lingual\\ \hline
          
       Claude\footnote{https://claude.ai/} & 2023.3 & 52 billion & - & Join the RLAIF training paradigm & Longer and more natural text editing than ChatGPT\\ \hline
          
       BlenderBot3\footnote{https://blenderbot.ai/} & 2022.8 & 175 billion & - & Instruction fine-tuning & Text generation, question answering\\ \hline
    \end{tabular}
    }
\end{table*}

\subsection{Fusion strategies}

\textbf{Cooperating with the education and training community.} LLM technology engages with schools, online education platforms, and educational technology companies to collectively explore and develop the application of LLMs in education. Partnering to provide actual educational scenarios and resources can help customize models to meet educational needs and accelerate the implementation of LLMEdu. For example, Baidu launched ``ERNIE Bot" \cite{zou2021pre}, Alibaba Group Holding Limited launched ``Tongyi Qianwen"\footnote{\href{https://qianwen.aliyun.com/}{aliyun.com} }, and universities like Tsinghua University launched "ChatGLM"\footnote{\href{https://chatglm.cn/main/detail}{ chatglm.cn} } \cite{zeng2023measuring}, etc.

\textbf{Form customized content generation to enhance competitiveness.} LLMs require high-quality and large data sets, so the education and training community can use LLMs to generate high-quality educational content, such as course materials, textbooks, exercises, and tests. For example, Baidu's ``ERNIE Bot" has a certain accuracy in answering knowledge questions because it uses the Baidu Encyclopedia as training material. ChatGPT can also generate some framework lesson plans for teaching.

\textbf{Provide popular educational functions.} Some educational technology companies develop an intelligent tutoring system, use LLMs to answer students' questions, provide answers and feedback, provide logical responses to open-ended questions, and provide guided responses to calculation questions. For example, MathGPT, developed by TAL, provides high-quality problem-solving tutoring in the field of mathematics \cite{scarlatos2023tree}. Some use LLMs to develop speech recognition and dialogue systems, making speech education and interaction easier to implement, enabling language teaching and situational dialogue \cite{kim2022improved}.

\textbf{Integrate LLMs into online education platforms.} Based on the learning model combined with the Internet and the rapid development of big data, integrating LLMs into online education platforms can provide students with richer learning resources, tools, and more comprehensive applications. For example, the Coursera online education platform\footnote{\href{https://www.coursera.org/}{Coursera | Degrees, Certificates, \& Free Online Courses} } uses LLMs to implement functions such as data collection and course recommendations. Duolingo\footnote{\href{https://www.duolingo.com/}{duolingo.com} } uses LLMs to upgrade language functions. Chegg\footnote{\href{https://www.chegg.com/}{Chegg - Get 24/7 Homework Help | Rent Textbooks}} uses LLMs to optimize the homework tutoring process.

\textbf{Participate in optimizing the educational work training process.} First, provide training and support to educators so that they can effectively use LLMs and related tools. For example, we learn how to integrate models into teaching, as well as how to interpret and use the data and recommendations generated by the models. Second, we use LLMs to analyze student data to provide educators with insights about student progress and needs, thereby optimizing their teaching methods, such as timely feedback features.

\textbf{Continuous improvement and research.} The gradual integration of LLMs into the education industry requires time and resources. During this process, the performance, application, and potential risks of LLMs are continuously monitored and improved, and data privacy and security regulations are observed, considering the educational needs of different regions and cultures, which can maximize the role of LLMs in the education industry.

\section{Key Technologies for LLMEdu}
\label{sec:ktfe}

\begin{figure}[ht]
    \centering
    \includegraphics[trim=0 0 0 0,clip,scale=0.125]{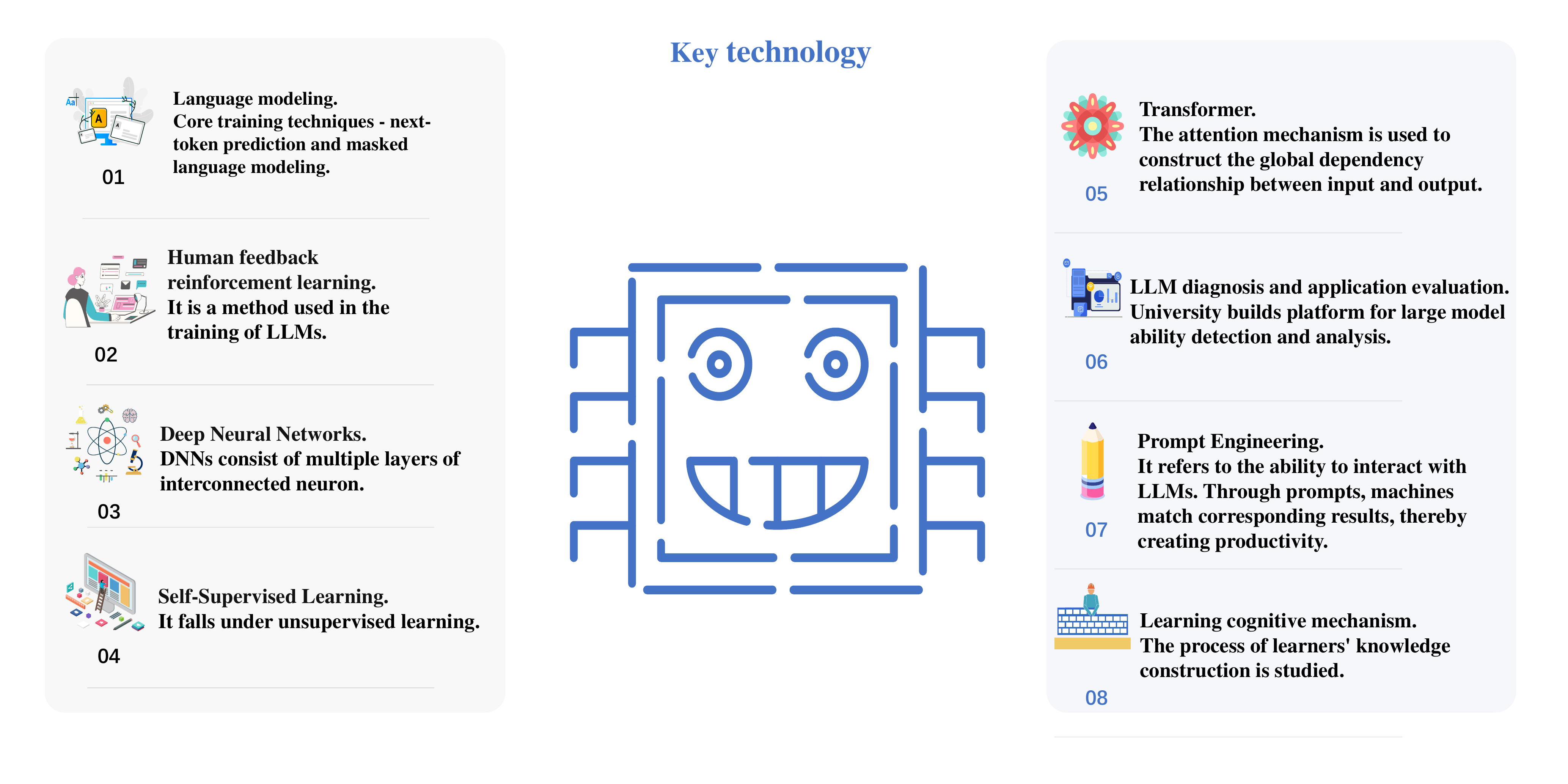}
    \caption{Key technologies of the LLMs }
    \label{fig:key}
\end{figure}

The technologies behind LLMs support their rapid development, as shown in Figure \ref{fig:key}. The combination of these technologies enables LLMs to achieve excellent performance in a variety of NLP tasks, such as text generation, machine translation, sentiment analysis, and text classification. They already play an important role in various applications such as virtual assistants, intelligent search, automatic summary generation, and natural language understanding, which promotes the development of LLMEdu.

\begin{table*}[!h]
\renewcommand{\arraystretch}{2.7}
    \centering
    \caption{Comparison between generative AI and discriminative AI}
    \label{DG}\resizebox{0.9\linewidth}{!}
    { 
    \begin{tabular}{|c|c|c|c|c|} \hline
       & \textbf{Core} & \textbf{Data learning}& \textbf{Development process} & \textbf{Application}\\ \hline
       \textbf{Discriminant/Analytical AI} & Analysis & Conditional probability distribution & Mature technology and widely used & Recommendation systems, CV, NLP\\ \hline
       \textbf{Generative AI} & Creation & Joint probability distribution & Exponential explosion & AIGC, text generation, audio generation\\ \hline      
    \end{tabular}
    }
\end{table*}

\textbf{Language model.} It learns from a corpus and predicts word sequences based on probability distributions. Two main technologies used to train a language model are next-token prediction and masked language modeling. Next-token prediction predicts the next word based on its context, and masked language modeling learns the statistical structure of language, like word order and usage patterns \cite{bao2020unilmv2,chung2021w2v,ng2023can}. However, there is still a significant gap between predicting text and mastering more advanced representations in LMs, so training strategies for LMs can be inconsistent and may not correctly reach the ultimate goal. The prediction ability reflects the large model's learning ability, which determines whether the LLM can form a coherent and logical text when answering questions. So the language model is LLMEdu's foundation.

\textbf{Human feedback reinforcement learning (HFRL).} It is a method used in the training of LLMs \cite{ouyang2022training}. By incorporating human feedback, it reduces distorted and meaningless outputs, helping ChatGPT overcome the issues present in GPT-3, such as consistency problems. It includes supervised fine-tuning, simulating human preferences, and proximal policy optimization \cite{zheng2023secrets}. i) In supervised fine-tuning, a small amount of annotated data is fine-tuned by first performing next-token prediction to improve the injected data, then integrating the results, and finally decoding operations \cite{fan2023grammargpt}. ii) Developing a reward model that simulates human preferences to rank the decoded results, and constructing a ranking sequence to obtain a scoring model. To ensure consistent annotation results, the ranking process uses ordinal ranking for data annotation, resulting in a new dataset composed of comparative data \cite{bakker2022fine}. iii) Proximal policy optimization aims to learn a policy that maximizes the cumulative reward obtained during training. The algorithm involves an actor, which outputs the probability distribution for the next action, and a critic, which estimates the expected cumulative reward for a given state. By iteratively optimizing the reward signal output, the model learns from experience, adapts to new situations, continuously adjusts its policy, and improves the LLMs \cite{wu2023pairwise}. HFRL improves LLMEdu's accuracy, making the output results more concise, accurate, and in line with the human thinking process.

\textbf{Deep neural networks (DNNs) \cite{guu2020retrieval}.} Before explaining DNNs, it is necessary to introduce deep learning. It refers to the learning of the underlying patterns and hierarchical representations of sample data, aiming to achieve the goal of machine learning with analytical capabilities similar to humans. DNNs consist of multiple layers of interconnected neurons, typically including an input layer, several hidden layers, and an output layer. The connectivity between neurons is similar to the connections between biological neural cells. DNNs have advantages in processing large-scale educational data, including students' academic performance, learning behavior, problem-solving abilities, etc. By analyzing these data, LLM can provide insights for educational decision-making and improve teaching methods and personalized education strategies.

\textbf{Self-supervised learning. } To produce the desired results, a model or machine needs to be trained with the given materials. Machine learning can be categorized into supervised learning, unsupervised learning, and reinforcement learning \cite{morales2022brief}. Self-supervised learning falls under unsupervised learning, where the model learns general feature representations for specific tasks. Unlike supervised learning, which requires a large amount of manually annotated data for training, self-supervised learning completes self-training by replacing human annotations with the intrinsic structural features of the data itself, using unlabeled datasets \cite{elnaggar2021prottrans,yan2022sam}. It gradually trains the parameters from scratch in a progressive manner, using part of the input as the supervisory signal and the rest as input. This approach significantly reduces the cost of manual annotation in terms of high cost, long cycles, and low accuracy, resulting in a lower development cost. Through self-supervised learning, LLMs can learn advanced representations of language data and deep cognition of language skills. This enables them to better understand and generate education-related content, including textbooks, exercises, solutions, and study materials.

\textbf{Transformer model.} From a structural perspective, LMs have evolved from statistical LMs to neural network LMs, and now to LLMs. Statistical LMs focus on transforming sentences into probability distributions, but the lack of computational power limits their ability to match massive amounts of data. Neural network LMs, such as recurrent neural networks, use recursion and convolutional neural networks to transform language sequences. Recurrent neural networks require considering the input-output order for computation and cannot handle examples in batches efficiently, resulting in slow speed. The Transformer model, widely used in LLMs, overcomes these limitations. The transformer model is essentially an encoder-decoder architecture that includes encoding and decoding components. It employs attention mechanisms to capture global dependencies between inputs and outputs \cite{derose2020attention}, without considering the distance within input or output sequences \cite{dong2023lambo}. This approach transforms the growth rate of required data for operations on related signals from linear or logarithmic to constant, showcasing high parallelism, which is beneficial for fast model iterations. Compared to previous models, the Transformer model has a richer structure, stronger adaptability to various scenarios, and better performance. The Transformer model improves the compatibility and practicality of LLMs, as well as its ability to cope with diverse and rich teaching contents and educational scenarios.

\textbf{LLM diagnostics and application evaluation.} Existing interdisciplinary evaluation systems assess LLMs from two perspectives: diagnostics during LLM training and the effectiveness of LLM applications. ``ChatbotArena"\footnote{\href{https://chatbotarena.com/}{24/7 Digital AI Assistant - Modern Chatbot (chatbotarena.com)} } is a benchmark platform for LLMs that conduct anonymous and random adversarial evaluations, where the system randomly selects two different LLMs to chat with users, who then rate the interactions. ``SuperCLUE"\footnote{\href{https://www.superclueai.com/}{SuperCLUE (superclueai.com)} } is a benchmark for evaluating general-purpose LMs in Chinese, examining multidimensional capabilities in terms of basic abilities, professional abilities, and Chinese-specific abilities \cite{xu2023superclue}. ``The C-Eval project" \cite{huang2023c}, jointly carried out by Shanghai Jiao Tong University, Tsinghua University, and the University of Edinburgh, constructs a multidisciplinary benchmark list to assist Chinese LLM research. ``FlagEval" \cite{li2023cleva}, built by multiple universities, adopts a three-dimensional approach to evaluating LLMs, including factuality, safety, and inclusivity. These evaluation frameworks are designed to comprehensively assess LLMEdu's performance, ethical impact, and potential bias, as well as promote the improvement of LLMEdu's capabilities and technology optimization.

\textbf{Prompt engineering \cite{naseem2021comprehensive}.} It refers to the ability to interact with LLMs. Machines match corresponding results through prompts, thereby increasing productivity. Good prompts can enhance the intelligence of LLMs and increase the value of feedback results \cite{tirumala2022memorization,zamfirescu2023johnny}, increasing the use value of LLMEdu. Moreover, poor prompts may lead to erroneous conclusions. In the field of education, especially rigorous science, the correctness of answers is always given priority, so optimizing prompt words is also important to deal with LLM's nonsense when answering academic questions. Different LMs, such as ChatGPT, ERNIE Bot, and MathGPT, have independent underlying training mechanisms, and their prompts are different. This can be likened to communication with individuals with different personalities.

\textbf{Learning cognitive mechanisms.} Learning cognitive mechanisms, which were developed in cognitive ethics, serve as the foundation for intelligent instructional design. It studies the process of knowledge construction in learners, integrating new knowledge into existing knowledge structures, and adjusting and updating the overall structure. Prior to ChatGPT, AI primarily focused on computation and reasoning. With AI's rapid development, its cognitive intelligence has gradually emerged and can even match human intelligence. There are two main cognitive approaches: one involves simulating human learning processes through computer models, and the other utilizes non-invasive brain imaging techniques such as functional magnetic resonance imaging. LLMs primarily simulate human learning processes, where pre-training can be likened to acquiring new knowledge and constructing knowledge.

By adding plug-ins, the latest LLM GPT-4 can address real-time problems, such as solving the lag problem of pre-training data. GPT-4 can also better solve logic problems because it introduces the mathematical problem data sets MATH and GSM-8K into the training data set, which greatly improves its mathematical reasoning capabilities. Moreover, GPT-4 can also complete creative text creation because it is connected to the API, and users can customize the AI character and complete simulated writing, reducing deviations and over-correction \cite{liu2023summary}.

\section{Implementation of LLMEdu} \label{sec:cases}

\begin{figure}[ht]
    \centering
    \includegraphics[scale=0.12]{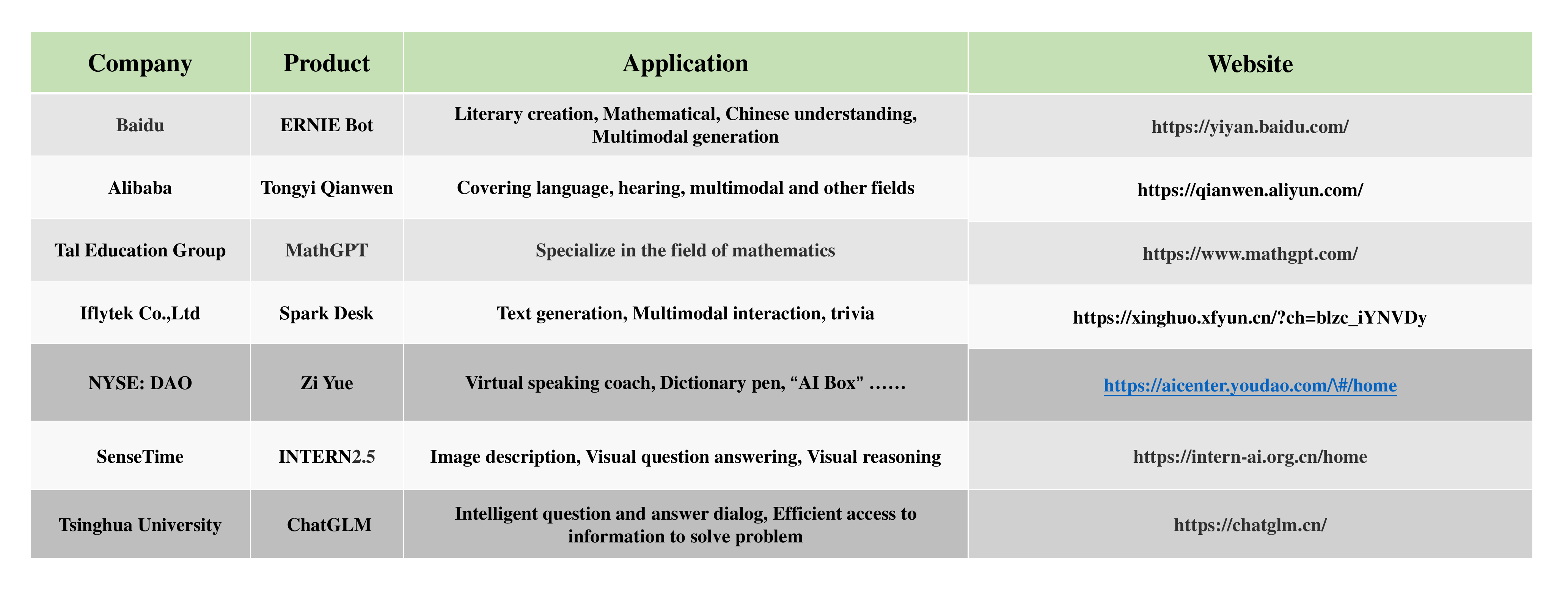}
    \caption{Examples of LLMEdu.}
    \label{fig:example}
\end{figure}

In this article, many products of LLMEdu are introduced, and the summary is shown in Figure \ref{fig:example}. Moreover, this part will focus on the implementation process of LMs from two aspects: LLMs empowering education and specifically LLMs empowering the field of mathematics. Finally, we use a unified framework to organize and compare the application of LLM in the field of education. The details are shown in Table \ref{application}.

\subsection{LLMs-empowered education}

\textbf{Improve teacher effectiveness.} LLM can help teachers access a wealth of teaching resources, allowing them to conduct classroom instruction more effectively. Before class, LLM can serve as a helpful assistant for lesson preparation. Through interactive question-and-answer sessions, LLM can provide ideas for teacher's lesson planning, assist in designing teaching outlines and curriculum plans, and help teachers quickly identify the highlights and challenges of a lesson. In the classroom, LLM can act as an AI teaching assistant, providing an instant feedback platform for both teachers and students and enhancing classroom engagement, interest, and appeal. After class, LLM can assist teachers in generating homework assignments and exam questions, enabling teachers to better assess students' understanding of the subject matter. In daily work, LLM is also a valuable assistant for teachers, capable of drafting meeting invitations, writing work plans, summaries, reports, and more. When used properly, LLM can help alleviate teachers' workload and promote their professional development \cite{zhang2023complete}. For example, a survey pointed out that during the paper revision process, 57.4\% of users believed that the feedback generated by LLM was helpful and could help them improve their research process \cite{Liang2023}.

\textbf{Promote student progress and growth.} In terms of learning assistance, LLM is a powerful tool that can understand complex concepts, solve difficult problems, and provide corresponding learning advice. In language learning, LLM offers scenario-based dialogue training, greatly enhancing student's oral and written abilities. In terms of cultivating thinking skills, LLM sometimes exhibits ``serious nonsense". Teachers and parents can utilize this phenomenon to cultivate students' critical thinking and enhance their information literacy. In terms of learning ability development, the process of using LLM requires students to ask questions. In this process, students have to learn how to translate their questions into effective questions and how to obtain useful information, which cultivates students' self-learning ability and summary ability. Taking college students as an example, data shows that more than 20\% of the users of one of LLM's latest products, the iFlytek Spark model, are college students, and it helps them improve in English speaking practice, mock interviews, and after-school homework.

\textbf{Answer professional and academic questions, accelerating research progress.} LLM is capable of writing academic experiment codes, building experimental models, quickly and accurately searching for literature materials, and extracting and integrating relevant information. This reduces the tedious process of manual research and accumulation, saving a significant amount of time. As a result, researchers can invest more energy into subsequent research, thereby improving research efficiency \cite{baidoo2023education}. Additionally, the report findings show that LLMs in universities, as an important research platform in the field of AI, have achieved remarkable results. Chinese universities’ research on LLMs mainly focuses on CV, NLP, speech recognition, and other fields. Research results in these fields not only provide a good academic atmosphere for teachers and students in universities but also provide strong support for the development of different AI industries.

\textbf{Promote the evolution of educational consciousness and form new learning paradigms.} The existing educational system is primarily focused on inheritance, and students often approach knowledge with inertial thinking inherited from their learning experiences. There is a lack of creative awareness. However, with the advancement of AI technologies such as ChatGPT, the existing learning paradigms are no longer sufficient for the future. Faced with the challenges posed by technologies like ChatGPT, it is necessary to cultivate higher consciousness and exercise thinking skills with a high level of awareness, forming new learning paradigms while improving perception and cognition to better understand the world. For example, the high-consciousness generative learning paradigm reflected in ChatGPT involves establishing connections between new and old knowledge, incorporating reflection and introspection, and innovating new concepts and understandings. To advance the high-consciousness generative learning paradigm, collaboration between educational designers and implementers is required to build adaptive learning environments and foster a positive learning atmosphere \cite{baidoo2023education}.

\textbf{Create highly contextualized and intelligent learning experiences.} In subject learning, generative AI like LLM, with its vast amount of data, can provide students with abundant information and knowledge, streamlining the process of finding learning materials and assisting students in finding answers and solving problems across various subjects. In language learning, LLM can offer real-time dialogue training, enabling students to immerse themselves in scenario-based learning and improve their conversational and writing skills. In terms of temporal and spatial aspects of learning, as an online tool, LLM can be accessed by students anytime and anywhere, providing great flexibility. Currently, LLMs are constantly improving their technologies and capabilities to achieve intelligent learning. For example, in the language understanding task, the ultra-large-scale Chinese pre-trained language model PLUG broke the Chinese GLUE classification list record with a score of 80.179. In the language generation task, it improved by an average of more than 8\% compared with the previous best results in multiple datasets.

\textbf{Promoting high-quality development in education enhances educational management and decision-making capabilities.} LLMs represent the latest technological means supporting intelligent education, and their development process reflects the synchronized progress of AI and humans. This embodies a new era of educational style that aims to create intelligence, cultivate wisdom, and create more efficient intelligence. Moreover, the data transparency involved in LLMs can make educational development decisions more precise and scientific, transforming educational decision-making from experiential patterns to evidence-based patterns and thereby enhancing educational governance capabilities. Finally, educational practitioners can use AI technologies like ChatGPT to conduct scenario-based assessments of students, resulting in a digital transformation of educational evaluation \cite{hawley2018student}. LLMs can help teachers judge student's progress in learning and understand student's learning status. Notice that the multi-dimensional data collected by LLMs through evaluation is helpful for educators to study student's learning logic and development rules, adjust teaching content on time, and provide students with personalized growth services.

\textbf{Driving in-depth research in the education system.} The research paradigms in education have evolved from the traditional observation and summary of scientific experiment experience, the construction of theoretical models and derivations, and computer simulation to the scientific research paradigm of large-scale data collection, analysis, and processing. The educational research paradigm is constantly changing. However, as time progresses, the old research paradigms no longer meet the requirements. The emergence of content-generative AI, represented by LLMs, has given rise to a new paradigm, "The Fifth Paradigm" of "AI for Science," enabling humans to delve further into the exploration of the education system. This paradigm shift involves the transition from simple imitation of humans to cognitive understanding and transformation, creating a new world of AI and education. According to a survey by Study.com\footnote{\href{https://study.com/}{Study.com | Take Online Courses. Earn College Credit. Research Schools, Degrees \& Careers} }, 21$\%$ of teachers outside China have begun to use ChatGPT to assist their teaching work. Chegg, a listed American education and training company, also said that after launching the LLM-based learning assistance platform, it has affected the user growth of its original business, and students' interest in ChatGPT has greatly increased.

\textbf{Promote the development of AI from fragmentation to scalability, thereby enhancing its generalization capabilities in education.} LLMs accurately capture knowledge from massive datasets through the process of pre-training an LLM and fine-tuning it for downstream tasks \cite{bender2021dangers}. This knowledge is stored in a large number of parameters and then fine-tuned for specific tasks. Finally, it can be flexibly applied to various scenarios. In other words, a single set of techniques can be used to address different tasks, greatly improving development efficiency. For example, in the field of education, LLMs share data to solve common problems and are widely applied in dialogue question-answering, language translation, text generation, and other scenarios. Some open-source LLMs, such as ChatGLM, Baichuan, InternLM, Qwen-7B, and Qwen-14B, are all manifestations of the generalization of LLMs, and Qwen-14B among them already has an accuracy of more than 70$\%$, which shows that these degrees are constantly improving.

\subsection{LLMs in Mathematics}
AI has been pursuing mathematical research and applications since its inception. Mathematics is a challenging subject in education, and proficiency in math represents a significant milestone in the intelligence level of LLMs. The successful handling of mathematical problems by LLMs will mark a new era in AI.

\textbf{Applications in mathematics can reflect the imitation ability of LLMs.} Mathematics is an abstract discipline that requires logical reasoning and critical thinking \cite{su2016mathematical}. Currently, LLMs are unable to genuinely comprehend the essence of mathematics and demonstrate independent thought. Therefore, when addressing mathematical problems, these LLM models rely heavily on the mathematical concepts and rules embedded in their training data. For instance, when solving algebraic problems, LLMs apply algebraic rules by mimicking the way humans learn and apply algebra \cite{liu2023summary}.

\textbf{Improvement of computational performance of LLMs in mathematics.} The essence of LLMs is to predict future outputs based on data correlation. However, errors may occur for symbols that are rarely or never encountered in the pre-training stage. For example, because the size of numbers is infinite and the scale of LLMs is limited, arithmetic operations on large numbers are likely to go wrong. To solve this problem, fine-tune the LLM on synthetic arithmetic problems and use special training and inference strategies to further improve numerical computing performance.

\textbf{Optimize the logical reasoning process.} One is to optimize the human logical reasoning process through LLMs. For example, some scholars have applied LLMs to the proof of theorems \cite{han2021proof}, because LLMs can provide a large amount of relevant materials to make up for the lack of information or omissions, making the reasoning more complete. The second goal is to improve LLMs' logical reasoning abilities. The logical reasoning ability of LLMs is a key indicator for evaluating LLMs. Because LLMs usually have problems such as excessive parameter space and severe data sparseness, LLMs perform poorly on robust and rigorous reasoning tasks. Relevant research has proposed optimization methods for LLM logical reasoning problems. For example, OpenAI\footnote{https://openai.com/research/improving-mathematical-reasoning-with-process-supervision} studies a process-based supervision model to improve the logical reasoning capabilities of GPT-4. Moreover, some research institutions use the method of continuous pre-prediction on large-scale mathematical corpora, which improves model performance on mathematical reasoning tasks.

\textbf{Interaction with external tools to improve LLMs' mathematical capabilities.} 1) LLMs interact with language conversion tools, such as lean language \cite{moura2021lean}, which can convert mathematical language into computer language, thereby improving the rigor of model reasoning. This is an innovative way to bridge the gap between human reasoning and machine reasoning. This could allow models to better understand and process complex mathematical concepts. 2) LLMs interact with information retrieval systems, such as the large dialogue model LaMDA proposed by Google, which connects to the information retrieval system and allows the model to learn to retrieve and use calculators and translation engines \cite{thoppilan2022lamda}. 3) LLMs directly interact with the calculation engine, such as MathGPT, which improves calculation accuracy by interacting with the calculation engine. This allows models to take advantage of calculators' powerful computing capabilities and perform complex mathematical calculations with greater accuracy. 4) LLMs enable themselves to determine the interactive tools, such as Meta's toolformer model, which can determine the use of external tools by itself \cite{Schick2023}. This gives models the flexibility to adapt to different situations and choose the most appropriate tools to solve a problem, much like humans do.

\textbf{Future development of LLMs in mathematics.} Specifically, the first is a cutting-edge exploration with scientific research at the core, such as the research and improvement of LLMs' capabilities in mathematics, including computing capabilities, reasoning capabilities, robustness, and so on. The second is to improve inclusive education and basic education for the general public. This entails studying how to use models to improve learning experiences and effects, as well as enhance mathematical education for students of all ages and backgrounds. By leveraging the power of LLMs, it may be possible to create personalized learning experiences that cater to individual student's needs and learning styles, making mathematics education more accessible and effective for a broader range of people. In terms of development potential, the expansion of LLMs' ability to solve mathematical problems could have far-reaching implications for other technical and educational fields. For example, LLMs could be used to improve the accuracy and efficiency of scientific simulations, enhance the effectiveness of machine learning algorithms, or even aid in the development of new technologies such as quantum computing. Ultimately, the development of LLMs in mathematics could drive the development of a new generation of education models that are more inclusive, effective, and efficient.

\begin{table*}[!h]
\renewcommand{\arraystretch}{1.815}
    \centering
    \caption{Comparison between generative AI and discriminative AI}
    \label{application}
    \resizebox{0.95\linewidth}{!}
    { 
    \begin{tabular}{|c|c|c|c|c|}
        \hline
        \textbf{Application} & \textbf{Advantage} & \textbf{Disadvantage}  & \textbf{Challenge} & \textbf{Future development} \\ \hline
       ~&Save time and costs&Data privacy issues&Expand the corpus&Develop personalized applications \\
   
        Personalized learning & Precise teaching & Information bias  & Information accuracy & Information extraction technology update\\
       
        ~& Good interactivity & The learning process is opaque & Update corpus in real time & Integration of various technologies \\ \hline
         
        ~&Improve problem-solving abilities & Marginalized teachers & Social impact & Training with more accurate data \\
           
        Guided learning &  Encourage critical thinking & Misleading information & Emotional understanding & Integrate with personalized experiences\\ 
        
        ~& Cultivate interest in learning & Lack of emotional resonance & Unemployment Risk & Develop policies to address social impacts \\ \hline
     
        ~& Provide diverse learning support &Insufficient training data support & Logic optimization&Integration of multidisciplinary and LLM \\
       
        Interdisciplinary learning &   Cultivate interdisciplinary thinking skills & Lack of domain knowledge & Accuracy of knowledge integration & Revolutionize the way we learn and teach\\ 
        
        ~& Boast excellent interdisciplinary capabilities & Disciplinary bias & Algorithm optimization & Filter useful training data \\ \hline
       
       ~& Reduce teacher stress & Machine hallucination & Multiple text associations & Standardize technology use \\
      
       Real-time problem-solving & Improved learning efficiency &Over-reliance on technology & Text extraction & Acceleration of model inference\\ 
       
        ~& Teaching assistance upgrade &~&~& Diversified technical assistance \\ \hline
       
        ~& Guide mathematics learning &~& Promote mathematical research & Pay attention to thinking guidance \\
           
       Applications in mathematics &   Improve math learning efficiency & Math terminology learning & Improved logical reasoning ability & Mathematics research and teaching\\ 
       
    ~& Show the fusion of AI and mathematics &~& Understand number relationships & Adequate language modeling \\ \hline        
    \end{tabular} }
\end{table*}

\section{Issues and Challenges} \label{sec:challenges}

\begin{figure}[ht]
    \centering
    \includegraphics[scale=0.128]{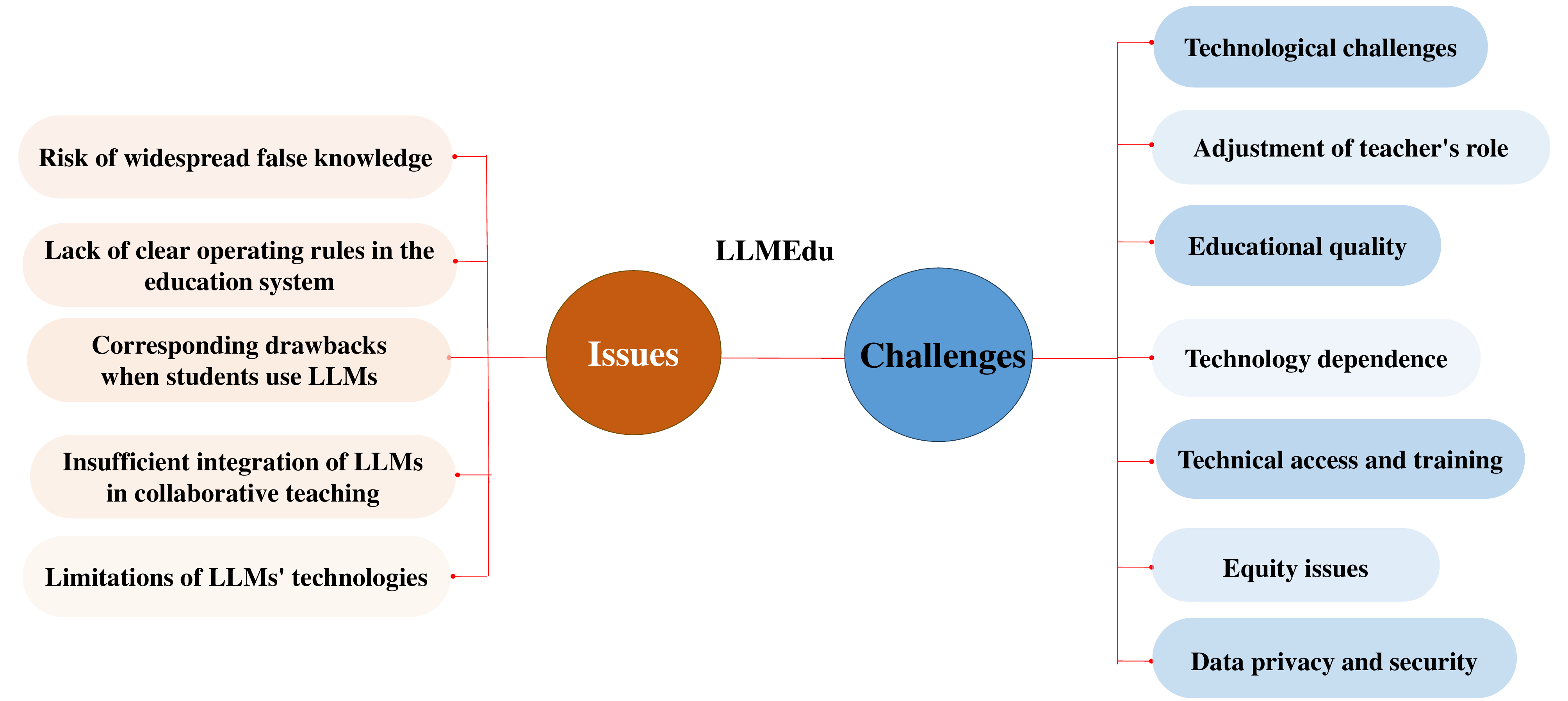}
    \caption{Some challenges and issues of LLMEdu.}
    \label{fig:question}
\end{figure}

In practical applications, LLMs for education still face many issues and challenges, including but not limited to, as shown in Figure \ref{fig:question}.

\subsection{Main issues}

\textbf{Risk of widespread false knowledge.} As an imperfect intelligent technology, LLMs such as ChatGPT still have many flaws. The biggest drawback is the potential for generating incorrect information \cite{amer2023large}. As many people have noticed, LLM sometimes exhibits machine hallucination \cite{rawte2023survey}. For example, a computer scientist in California tried different methods to check the output of the GPT robots and found that GPT-3.5 and GPT-4 were full of errors when testing physics, chemistry, and mathematics questions selected from college textbooks and exams. Moreover, since LLM's training data largely consists of English corpora, it often struggles to understand and provide correct answers to personalized Chinese questions. In the short term, these errors can cause disruptions in students' knowledge learning, and students with weaker discernment abilities are highly likely to acquire erroneous knowledge without realizing it. In the long term, if the corresponding technology is not improved promptly, LLM may contribute further to the proliferation of false knowledge. There are many examples of actively dealing with machine hallucinations. For example, the retrieval-augmented generation method (RAG) can integrate LLM with a rigorously verified external key knowledge corpus.

\textbf{Lack of clear operating rules in the education system.} Due to the complexity of education itself, representing the education system using specific symbols and algorithms is an extremely challenging process that current LLMs cannot achieve. Education behaviors, such as emotional interaction, effective communication, and leading by example, are currently beyond the capabilities of LLMs. LLMs learn from a large amount of data and provide feedback, representing subjective educational information with data and providing rational reflections of human thinking. The goal of anthropomorphizing LLMs is to enable NLP models, such as Word2Vec, to convert words into vectors, facilitating the computer's processing of textual data \cite{amin2023will}. GPT-1 and BERT, based on the self-attention mechanism \cite{ghojogh2020attention}, further enhance performance. GPT-3 achieves another leap in performance on zero-shot learning tasks with its significantly increased parameter scale \cite{Wei2021}. ChatGPT's HFRL, code pretraining, and instruction fine-tuning improve the model's inference capabilities \cite{ouyang2022training}. GPT-4, an ultra-large-scale multimodal pre-trained model, possesses multimodal understanding and multi-type content generation capabilities \cite{LILI2023}. These examples show ideas for solving the problem of anthropomorphizing LLMs, gradually approaching human-like capabilities through continuous optimization and development, thereby alleviating the limitations of the abstraction and ambiguity of educational rules.

\textbf{Some drawbacks when students use LLMs.} The occasional inaccuracies in LLM's answers can mislead students who lack critical thinking skills. The great convenience of LLM may reduce students' desire for independent learning and innovation, leading to intellectual laziness. As LLM involves massive amounts of data, students who lack awareness of data security may unknowingly leak their personal data \cite{yu2023codeipprompt}. While LLM provides interactive dialogue scenarios and opportunities for AI communication with students, it reduces real interpersonal conversations, and the way of discussing problems may shift from online to one-sided questioning of the machine, affecting the development of student's social skills. In response to these problems, educators need to actively guide students to adapt to the characteristics of LLM-assisted education and enhance the cultivation of privacy and security awareness.

\textbf{Insufficient integration of LLMs in collaborative teaching \cite{liu2023summary}. }Although LLM has achieved some level of one-on-one dialogue and communication, its integration with education in real life is still limited. The ability to solve higher-order reasoning problems and complex problems still needs improvement. For example, while GPT-4 performs reasonably well in some exams, it fails to demonstrate significant advantages in logical reasoning problems \cite{LiuLIU2023}. Most LLMs have high accuracy rates (up to 95\%) for reasoning with a small number of steps, but as the number of steps increases, reaching 20 or more, the accuracy drops significantly to 36\%, indicating a significant disparity \cite{Paranjape2023}. As a result, it is necessary to develop chain-of-thought technology to improve LLMs' reasoning ability and ability to solve complex problems \cite{wei2022chain}, thereby promoting the integration of large models and collaborative education.

\textbf{Limitations of LLMs \cite{thirunavukarasu2023large}.} Firstly, in pre-training, models that simultaneously satisfy the reasonable model size, advanced few-shot learning capability, and advanced fine-tuning capability have not been achieved yet. For example, GPT-3 lacks a reasonable model size and is relatively large in scale \cite{Brown2020}. Furthermore, the high complexity and strong data dependency of LLMs may be exploited by malicious data to affect their training process and generation results, as well as output uncertainty and other factors. The lack of interpretability in LLMs' technology makes their internal mechanisms unclear. The widespread application of LMs requires interpretability to ensure application security, overcome performance limitations, and control societal impact, which has triggered corresponding considerations regarding these issues. In the future, LLM's technology still needs optimization and innovation, and researchers need to consider the interpretability of the model more based on the user's situation.

\subsection{Main challenges}

\textbf{Technological challenges.} The application of LLMEdu relies on AI-based technologies, which are complex and challenging. If the technology is not perfected, it becomes difficult to provide high-quality educational services. The availability of high-quality data sources is one important factor influencing the improvement of LLM technology. High-quality data transformation involves capture and conversion processes. It is necessary to consider how to expand the perception of the educational field to capture dynamic performance data from any learning activity in educational subjects and how to improve the quality of the data through efficient processing. Moreover, LLMEdu faces technological challenges such as speech recognition, NLP, AIGC \cite{wu2023ai}, multimodal LLMs \cite{wu2023multimodal}, and other aspects. The above-mentioned issues require researchers to always pay attention to the development of other technologies in the AI field and actively integrate them into LLM to bring a better experience to the education industry.

\textbf{Artificial intelligence security.} The intelligence level of LLMs continues to improve, and security issues have become more severe. The first is the LLMs' biased cognition. Some studies have pointed out that when LLMs are tested using gender bias data sets, their answers will reflect gender bias \cite{kotek2023gender}. Therefore, when training an LLM, the data should be filtered. The second is the lack of correct social, moral, and ethical values. For some issues that violate social ethics, LLMs are unable to judge, which increases the risk of crime. Therefore, the country should formulate a more complete legal system to regulate the use of LLMs. The third is the most common issue among artificial intelligence ethical issues: "AI replaces human activities". AI has limitations in education. While AI has great potential in education, it cannot replace the role of teachers, such as encouraging critical thinking, solving complex problems, and providing psychological and social support. However, humans should also flexibly adjust their roles, regulate and guide the development of AI from an ethical perspective, and maintain their dominant position.

\textbf{Education quality.} The use of LLMEdu provides many opportunities for smart education, but it also presents challenges in terms of quality. If LLMEdu cannot provide high-quality educational services, it will be difficult to gain recognition from students and teachers. Furthermore, educational institutions that use LMs must strike a balance between educational quality and technological innovation. Otherwise, there may be an overreliance on technology, neglecting the quality of education itself. Therefore, to ensure the quality of education, the first consideration is to ensure the educational content, which requires educators to adjust reasonable teaching content and clarify the auxiliary functions of LLMs. Then, technology developers are required to ensure that the technology of LLMs is steadily progressing.

\textbf{Technological dependence. } Note that the future LLMEdu should be human-centric but not technology-centric \cite{yang2023human}. Overreliance on AI may reduce students' ability for independent learning and innovative thinking, and it may even lead to cheating and academic misconduct, such as using ChatGPT to complete assignments and papers. It is necessary to prevent the passive application of LLMs, as seen in the examples in reality. While using AI, the student should be encouraged to think independently, explore problems, and find answers. Furthermore, students should be educated on time management, ensuring sufficient time for other important activities while using AI, and avoiding excessive dependence on it.

\textbf{Technical accessibility and training.} The introduction of AI technology requires corresponding hardware infrastructure and network support. In resource-limited areas, this can be a challenge. Combined with the pressures and entrenched thinking that fear is being replaced \cite{yan2023practical}, there is a phenomenon of fear and refusal to use AI in education, in other words, cognitive limitations. In such cases, technical access and training become difficult. Therefore, efforts should be made to promote the long-term advantages of AI in the education industry, guide teachers and students to receive appropriate training, better understand the application ideas and specific methods of intelligent technology, enhance willingness to use, and better adapt to and utilize these tools.

\textbf{Equity issues.} Although AI has the potential to improve the quality and efficiency of education, its use can lead to unfairness among students. For example, some families may not be able to afford AI learning tools, or in certain areas, students may lack access to the necessary technological facilities for tools like ChatGPT. Educational equity is the cornerstone of social development, and interventions are needed to address the examples mentioned above effectively. For instance, when designing and optimizing LLMs, efforts should be made to balance characteristics such as race, gender, and age, reducing the digital divide and gender gap.

\textbf{Data privacy and security \cite{yu2023codeipprompt}.} Data privacy, including privacy protection, is a significant concern in the application of LLMs. LLMs involve collecting personal information and learning data from students and teachers. Therefore, privacy protection becomes an important issue in LLM applications. Educational institutions need to ensure the effective protection of student's and teacher's privacy while also ensuring the security and reliability of the data. Parents and teachers should focus on cultivating children's awareness of data privacy and security, as well as educating students to avoid privacy risks associated with the use of LLMs. Moreover, when collecting and processing student's learning data, it is essential to ensure that this information is properly protected to avoid data breaches or improper use.

In the future, following the development characteristics of the era of integrating intelligence and education, while continuing to optimize core technologies and technological innovations, LLMs such as ChatGPT, GPT-4, and MathGPT will continue to empower the education field. Moreover, based on the existing LLMs, we must continue to look for more effective training methods to more efficiently train models with large-scale parameters \cite{bender2021dangers}.

\section{Conclusion} \label{sec:conclusion}

In this article, we have introduced the development and application of LLMs in the field of education as comprehensively as possible. There are still some technologies that have not been included, as well as other issues that have not been discussed in depth. It is hoped that the technology introduced in this article and the thinking presented can help scholars and researchers better develop and optimize educational LLMs. This article summarizes the process of integrating education and LLMs. LLMs have excellent language generation and interactive capabilities that cannot be provided by traditional book-based teaching. It demonstrates the creative role of AI in education, as well as teachers, and the changing roles of parents and students. For smart education, we call for more mature education and AI development standards, technical specifications, and data security guidelines to focus on more practical issues. How to ensure data security? How can we limit the behavior that relies too much on AI technology? How to cultivate students' active exploration abilities?
LLMs and education complement each other. The application of LLMs in education makes education more intelligent and efficient, and the data accumulated over many years in education can help optimize LLM training. More attention should be paid to these development conditions. How can we create more valuable LLMEdu application scenarios? We look forward to the future of LLMEdu.

~\\

\textbf{Acknowledgments} This research was supported in part by the National Natural Science Foundation of China (No. 62272196), the Natural Science Foundation of Guangdong Province (No. 2022A1515011861), Guangzhou Basic and Applied Basic Research Foundation (No. 2024A04J9971). 

\textbf{Author contributions} Hanyi Xu: paper reading and review, writing original draft. Wensheng Gan: conceptualization, review and editing, supervisor. Zhenlian Qi: conceptualization, review and editing. Jiayang Wu: writing original draft. Philip S. Yu: review and editing.

\textbf{Data availability} This is a review paper, and no data was generated during the study.

\textbf{Conflict of interest} The authors declare that they have no known competing financial interests or personal relationships that could have appeared to influence the work reported in this paper.

\printcredits

\bibliographystyle{cas-model2-names}

\bibliography{LLMEdu.bib}

\end{document}